# Characterizing how 'distributional' NLP corpora distance metrics are

Samuel Ackerman*     George Kour†     Eitan Farchi†

**Abstract**
A corpus of vector-embedded text documents has some empirical distribution. Given two corpora, we want to calculate a single metric of distance (e.g., Mauve, Frechet Inception) between them. We describe an abstract quality, called 'distributionality', of such metrics. A non-distributional metric tends to use very local measurements, or uses global measurements in a way that does not fully reflect the distributions' true distance. For example, if individual pairwise nearest-neighbor distances are low, it may judge the two corpora to have low distance, even if their two distributions are in fact far from each other. A more distributional metric will, in contrast, better capture the distributions' overall distance. We quantify this quality by constructing a Known-Similarity Corpora set from two paraphrase corpora and calculating the distance between paired corpora from it. The distances' trend shape as set element separation increases should quantify the distributionality of the metric. We propose that Average Hausdorff Distance and energy distance between corpora are representative examples of non-distributional and distributional distance metrics, to which other metrics can be compared, to evaluate how distributional they are.

## 1. Introduction

Generative models have garnered significant attention due to their capacity to generate novel data points that resemble real-world examples. However, the efficacy of generative models hinges crucially on the evaluation process. Unlike classical, largely discrimination-concerned, machine learning, which predominantly focuses on classification accuracy, the evaluation in generative models is a more intricate task [13]. It necessitates a multifaceted approach, encompassing measures such as likelihood scores, quality of generated samples, and the diversity of generated data. In terms of diversity, it's crucial to capture how closely the generated data's distribution matches that of the training data. While applications to generative text models motivate this work, our experiments here will not involve text generation but only the models' text embedding abilities. This ensures a robust assessment of the generative model's capabilities.

Let $a_i$ denote a textual string or document[1]; here we will assume they are sentence-length. A corpus consists of a set of such documents. Many metrics, such as cosine distance, exist for determining how similar two such documents are. Here, we consider instead two corpora, say $A = [a_1, \ldots, a_{|A|}]$ and $B = [b_1, \ldots, b_{|B|}]$. If we imagine that these corpora represent random samples from two unknown distributions, that is, $A \sim \mathcal{P}_A$ and $B \sim \mathcal{P}_B$, we may be interested in characterizing the overall distribution or sample distance—denoted $d(A, B)$, where $d$ is a given distance metric function—between the pair of corpora, as opposed to the distances between individual component documents. Such metrics are of interest in many applications, such as text generation, where $A$ may be some source corpus, and $B$ is generated by some algorithm to be similar to $A$, and we may want to validate this generation algorithm by verifying that $d(A, B)$ is small, for instance.

---

*IBM Research, Haifa; samuel.ackerman@ibm.com
†IBM Research, Haifa

[1]The approach detailed in this work may also apply to $a_i$ of other domain types, such as images, because the method does not assume the representation must be of text. However, our experiments are restricted to the textual domain.

In prior work [6], we presented a collection of such corpus distance metrics $d(\cdot, \cdot)$, and proposed a framework to evaluate the behavior of these metrics according to desired criteria. For instance, the calculated $d$ should increase monotonically (perhaps linearly) when corpora samples $c_i \sim \mathcal{P}$ and $c_j \sim \mathcal{Q}$ are drawn from distributions known (without reference to $d$) to be more dissimilar; we used the term 'measures' for these properties of distance metrics.

Here, we wish to characterize an additional measure, or aspect of distance metrics, called the metric's 'distributionality'. This aspect relates more to how the metric works internally than do the other measures, which capture performance or discriminability. A distance metric that is more distributional considers the overall distributions' shapes when calculating the inter-corpora distances; a relatively non-distributional metric reflects more local measurements (e.g., distances to nearest neighbors), even if it takes into account global measurements. The crucial aspect is that a non-distributional metric should be more sensitive to small local changes—for instance, if it uses infimums of global measurements—that do not much affect the distribution shape overall. While we can know inherently or experimentally how global the considerations are, we would like to measure this quantitatively.

Section 2 shows synthetic numeric simulations of the concept of distributionality in in a non-text-based scenario. Section 3 reviews the relevant background on Known-Similarity Corpora from [6] that are necessary for this discussion. Section 4 introduce the concept of paraphrase corpora used in our experiments. Section 5 shows results of our attempt to measure the distributionality of corpus distance metrics. Section 6 concludes. Code for the paper experiments is available at https://github.com/IBM/text-corpus-distance-distributionality.

## 2. Illustration of distributionality

Before discussing the modeling of the distributions of text corpora and the distances between them, we first illustrate the concept of distributionality on simple synthetic numeric data. Say we have samples $A = [a_1, \ldots, a_m]$ and $B = [b_1, \ldots, b_m]$ of equal size $m$, where each $a_i$ and $b_i$ are $\in \mathbb{R}^q$ for some dimension $q \in 1, 2, \ldots$. Ultimately, we are interested in text paraphrase corpora (see Section 4) in which there is some pairing relation between elements $a_i \in A$ and $b_i \in B, \forall i = 1, \ldots, m$. We use notation $a_i \sim b_i$ to show these two elements are paired, which means the tend to be closer in distance to each other than to other elements. Two samples $A$ and $B$ where elements are all independent and identically-distributed (IID) will not typically display such a pairing between elements, hence we do not make this assumption.

In this illustration we generate an initial sample $A$ where $a_i$ are IID from a $q$-dimensional mixture Gaussian distribution, simulated using `sckikit-learn`'s ([8]) `datasets.make_blobs` function, where each of 3 mixture component has distribution $\mathcal{N}(\mu_j, \kappa \boldsymbol{I}_q)$ around its centroid $\mu_j$, $j \in \{1, 2, 3\}$; the centroids are distributed uniformly within the space by `make_blobs`. Then, each $b_i$ is generated by jittering its paired value $b_i \sim \mathcal{N}(a_i, \sigma \boldsymbol{I}_q)$, where $\sigma \ll \kappa$, so that the pairs $(a_i, b_i)$ are closer than average pairs in the component. The left plot of Figure 1 shows an example of such paired samples where $A$ is the solid dots and $B$ is the hollow circles.

In prior work, we introduced several inter-corpus distance metrics $d$. The ones discussed here are Fréchet Inception Distance (FID), Mauve, Precision-Recall (PR), Density-Coverage (DC), 'Classifier' (F-1 score of corpus separation by SVM model), and Information-Retrieval Precision-Recall (IRPR). See [6] for full details. In this present work, we see to analyze the distributionality aspect of these metrics. We propose that non-distributional metrics are more sensitive than distributional ones to small local changes. This should be true of the IRPR metric, and DC and PR when the value of $k$ (the number of nearest-neighbors used) is low, since a small change can affect the $k$-nearest-neighborhood of more points than if $k$ is higher. For instance, if a distance metric between $A$ and $B$ depends on how many elements of $y \in B$ are $k$-nearest neighbors of each element $x \in A$, perturbing a group of points together can affect this metric.

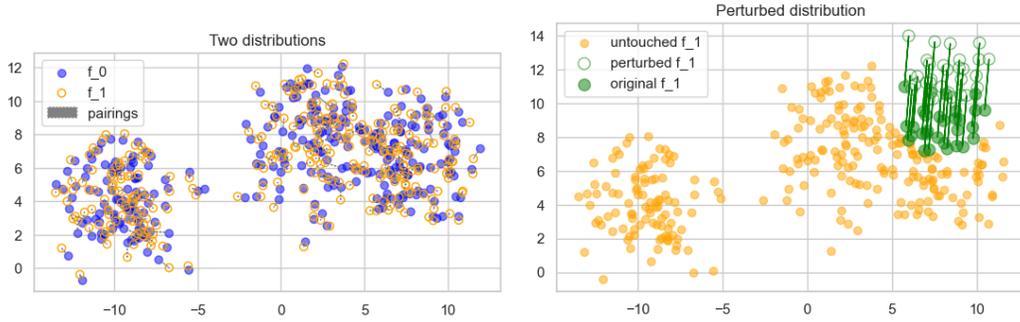

**Figure 1**: Left: Randomly-generated sample $A$ of size $m = 300$ (blue solid dots), with paired sample $B$ (hollow yellow dots) jittered from each element $a_i \in A$. Right: Example of shifting perturbation of a group of 10% of points (solid green dots) in $B$ in a fixed direction away from the centroid, by distance $\delta = 3$ (hollow green dots).

The right plot of Figure 1 shows an example of a random local perturbation of a proportion $p = 0.10$ of points of $B$ from the left plot. The perturbation operates by first randomly drawing an origin point $y^* \in B$. Then, for $k = \text{round}(p \times |B|)$, the $k$-nearest-neighbors of $y^*$ (including itself), denoted $NN_k(y^*) \subseteq B$, are moved together in the same direction by a distance of some $\delta > 0$. The direction of movement can be random[2] or forced to be in the direction of the vector $z = (y^* - \text{centroid}(y^*))$, which pushes the points away from the centroid that $y^*$ comes from; our experiments use the second option.

We now illustrate the results of applying perturbations to the paired distributions $A$ and $B$ shown in the left plot of Figure 1. In Figure 2, the proportion of points perturbed is fixed at $p = 0.1$ and the perturbation distance $\delta$ increases gradually. In Figure 3, the perturbation distance is fixed at $\delta = 3$ and the proportion $p$ increases gradually. Both perturbations designed to affect the global distribution shape but in different ways, and thus illustrate the distributionality aspect of the distance metrics; the important part is that only one perturbation parameter is allowed to increase. Each time, the perturbation is repeated $r = 20$ times on a different randomly selected origin point $y^* \in B$, generating a perturbed sample $B'$. Various distance metrics $d$ from [6] are applied to measure $d(A, B')$, which are pooled across repetitions. The CHISQUARE and ZIPF metrics are omitted in this illustration because they require integer-valued frequency vectors rather than continuous valued ones. Furthermore, for each of the DC and PR metrics, we try $DC_k$ and $PR_k$, $k = 2, 5, 15$, where $k$ determines the nearest-neighborhood size to determine for each point when calculating the distance (not the same as the $k$ used for the perturbation); in [6], the default was $k = 5$, so $DC_5$ and $PR_5$ correspond to the metric used in [6].

---

[2]This is done by drawing a point $z$ from the surface of a $q$-dimensional sphere with radius $\delta$ (i.e., the vector $z$ has magnitude $\delta$) and adding it to each point in $NN_k(y^*)$. This operation preserves the same distance shift regardless of dimension $q$, while adding random Gaussian noise of a fixed covariance does not.

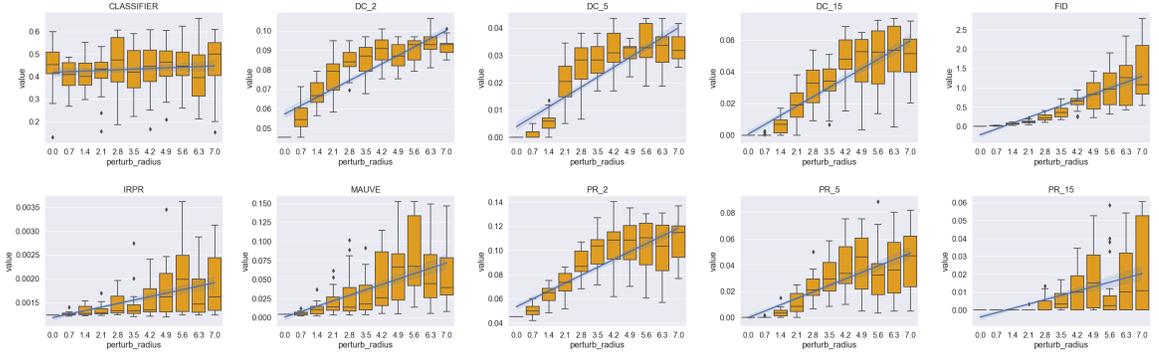

**Figure 2**: Distributions of distances $d(A, B')$ where $B'$ is perturbed $r = 20$ times from $B$, with fixed proportion $p = 0.1$ and varying perturbation distance $\delta$.

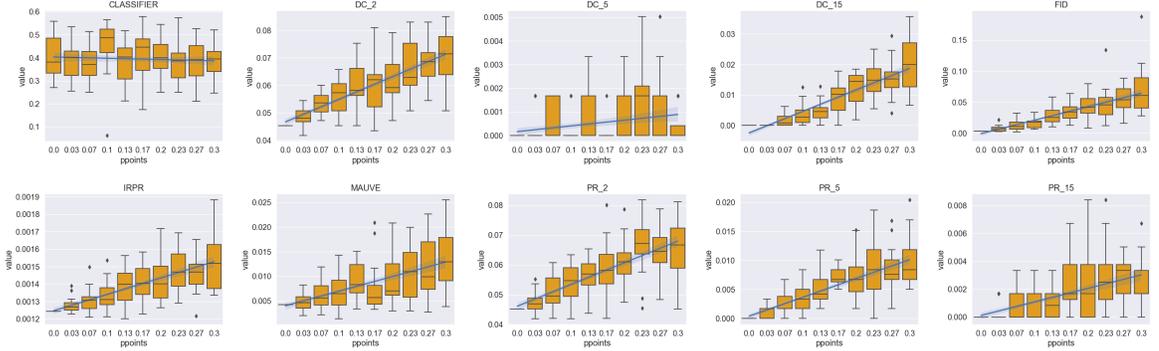

**Figure 3**: Same as Figure 2, except the perturbation distance $\delta = 3$ is fixed and the proportion $p$ is varied.

In general, a distance metric $d$ that is relatively non-distributional should be more sensitive to small local changes (here simulated by the group shift of points in the same direction). This means that $d(A, B')$ should tend to increase even at small values of the perturbation parameters $p$ or $\delta$; a relatively distributional metric should tend to be less sensitive, meaning the distance should be low (similar to $d(A, B)$ with no perturbation) initially along the x-axis, and then begin increasing later. The simulations above, in particular Figure 2, confirm our intuition that IRPR is relatively non-distributional. Its distances increase fairly consistently with the increases in $\delta$. Similarly, at low $k$, the $DC_k$ and $PR_k$ distances increase faster while at higher $k$ they remain low initially and increase later. This confirms our intuition that when $k$ is lower, these metrics are more sensitive to these small changes, and thus less distributional. In contrast, the FID metric—which uses mean and covariance matrix estimates reflecting the overall sample distribution shapes, and should therefore be more distributional—shows a much slower initial increase as perturbation increases.

Figure 4 shows results when the feature dimension $q$ increases. Here, for each value of $q$, a single new pair of $q$-dimensional samples $(A, B)$ are generated once, and the random perturbations repeated $r = 20$ times. Although we do not include this in our criterion of distributionality, it appears that the more distributional metrics ($DC_k$ and $PR_k$ for higher $k$, FID, and Mauve) may also be less sensitive to changes in dimension $q$ for a given perturbation.

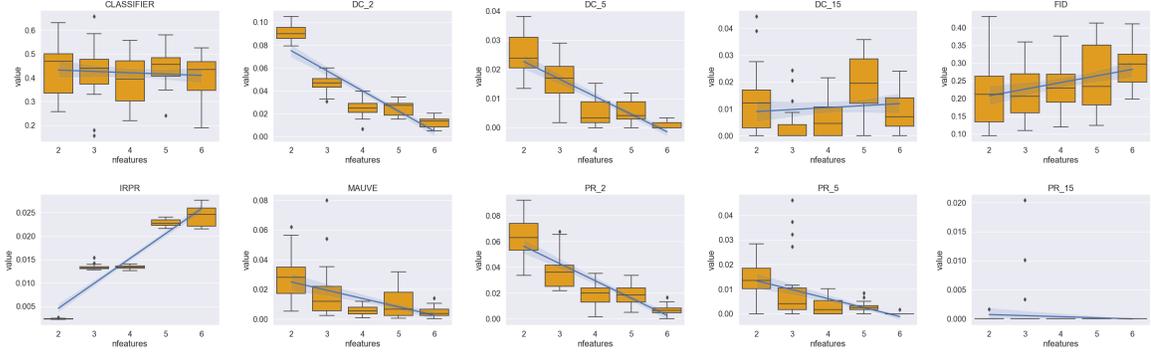

**Figure 4**: Same as Figure 2, except $p = 0.1$ and $\delta = 3$ are both fixed, and the dimension $q$ increases from 2–6.

## 3. Known-Similarity Corpora

In [6], measures were formulated by constructing a Known-Similarity Corpora (KSC) set, a framework developed by [5]. Given two source corpora $A = [a_1, \ldots, a_{|A|}]$ and $B = [y_1, \ldots, y_{|B|}]$, and an integer $K \geq 1$, a KSC set $KSC(A, B)$ consists of $K+1$ corpora[3] $\{c_0, \ldots, c_K\}$, each consisting of $n \geq K + 1$ documents. Let $p_i = i/K \in [0, 1]$. Corpus $c_i$ is generated by sampling $n_B(i) = \text{round}(np_i)$ observations from $B$, and the remaining $n_A(i) = n - n_B(i)$ from $A$; thus, $c_0$ and $c_K$ are sampled entirely from $A$ and $B$, respectively. We call this sampling procedure for each $c_i$ a **double lottery**, because each corpus $c_i$, $i = 0, \ldots, K$ is formed by concatenating two independent discrete uniform samples of specific sizes from the source corpora $A$ and $B$. The set $KSC(A, B)$ thus comprises a set of double-lottery draws of a mixture distribution of $A$ and $B$ where the mixture proportion $p_i$ increases from 0 to 1 at equal intervals of $\frac{1}{K}$. Figure 5 shows an illustration of $KSC(A, B)$ of length $K + 1 = 6$.

For a positive integer $a$, let the notation $[a] = \{0, \ldots, a\}$. Given $KSC(A, B)$ we then defined, for each $\ell \in \{1, \ldots, K\}$, the $\ell$-**distant corpora set** as

$$KSC_\ell(A, B) = \{(c_i, c_j) \colon i, j \in [K], j - i = \ell\} \qquad (1)$$

$KSC_\ell(A, B)$ consists of all ordered pairs of corpora $c_i, c_j \in KSC(A, B)$ where $j = i + \ell$, that is $c_j$ is $\ell$ 'separation steps' from $c_i$ in the direction of $B$. For instance, $KSC_1(A, B) = \{(c_0, c_1), (c_1, c_2), \ldots, (c_{K-1}, c_K)\}$, and $KSC_K(A, B) = \{(c_0, c_K)\}$ only. Furthermore, we defined

$$D_\ell(A, B, d) = \{d(c_i, c_j) \colon (c_i, c_j) \in KSC_\ell(A, B)\} \qquad (2)$$

as the set of values, given a corpus distance metric $d$, for all pairs in a given $KSC_\ell(A, B)$. The logic of the KSC set is that the sampling proportions $\{p_i\}$ governs the expected difference between two sample corpora. In particular, for two separation levels $\ell_1 < \ell_2$, we expect that on average, for $d_1 \in D_{\ell_1}(A, B, d)$ and $d_2 \in D_{\ell_2}(A, B, d)$, that $d_2 > d_1$, that is $E(I(d_2 > d_1)) > 0.5$. The expectation $E$ is calculated empirically over both sampled pairs in $KSC_{\ell_1}(A, B)$ and $KSC_{\ell_2}(A, B)$, and over repeated random generations of $KSC(A, B)$. Furthermore, values in $D_{\ell_1}(A, B, d)$ and $D_{\ell_2}(A, B, d)$ should be internally similar. This means that the separation level $\ell$ should be discriminative of distances $d$. In [6] the evaluation measures were based on relating changes in $\ell$ to $D_\ell(A, B, d)$, for instance how linear or monotonically increasing the relationship is.

---
[3]Note, here the starting corpus index is 0 rather than 1, as in that paper.

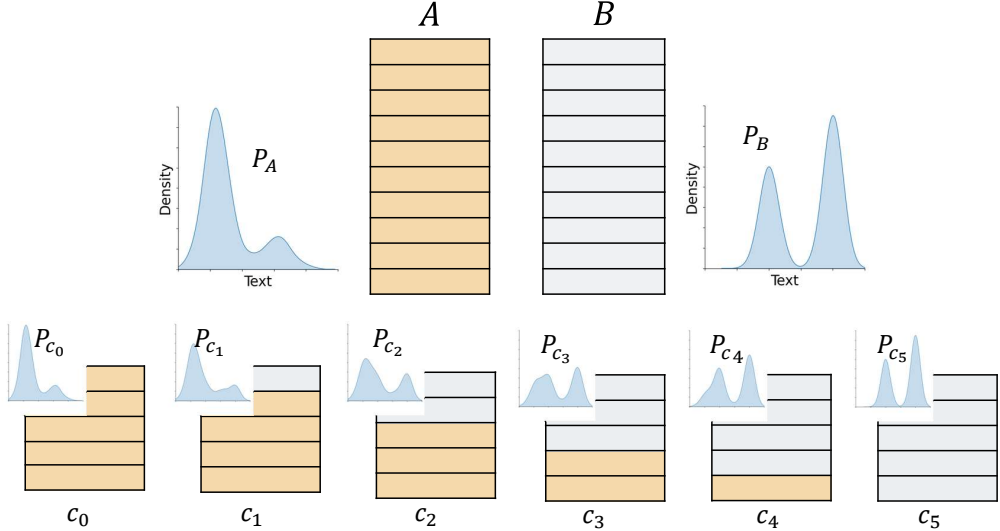

**Figure 5**: Illustration of a KSC set between two source corpora $A$ and $B$. Each shaded rectangle in each corpus $c_i$ corresponds to a fixed proportion $1/K$ (when $n = K$, to a specific document) that is sampled from either source corpus $A$ (yellow) or $B$ (grey).

## 4. Paraphrase KSC corpora

The experiments in [6] were based on source corpora $A$ and $B$ from different domains (e.g., medicine vs entertainment datasets) and are thus expected to contain documents that are semantically different. Here instead we consider $A$ and $B$ that are paraphrase corpora. In the following experiments we use the training split of the Quora question pairs dataset [9], which consists of 404,290 question pairs from the Quora online forums website, and is used in paraphrase classification problems. Of these, 149,263 pairs ($\approx 37\%$) are paraphrases (decided by human annotations), and we restrict our attention to this portion.

Note we will use the notation $d$ to refer only to inter-corpus distance metrics, and $\delta$ only to distances between individual documents (i.e., members of a corpus). A distance metric $d(A, B)$ may be based on measured values of $\delta$ between documents in $A$ or $B$. Furthermore, all distance metrics $d$ discussed here operate on the documents after being projected to an embedding space; this vector projection of a document $a$ will be denoted $\mathcal{V}(a)$. We use the embedding model `all-MiniLM-L6-v2` from Python's `SentenceTransformers` ([10]). For conciseness, we omit $\mathcal{V}$ in the notation. Thus, for instance, $\delta(a_i, b_j)$ is shorthand for the document-level distance $\delta(\mathcal{V}(a_i), \mathcal{V}(b_j))$, after each document is embedded through $\mathcal{V}$; the corpus-level distance $d(c_i, c_j)$ likewise is shorthand for $d(\mathcal{V}(c_i), \mathcal{V}(c_j))$, after the same embedding is carried out on each document in the corpora.

Figure 6 shows examples of rows of the Quora dataset. Each row consists of pairs of questions are paraphrases of each other, meaning they have essentially the same semantic meaning but use different words or structure, and thus are not identical. Corpora $A$ and $B$ are thus one column each of the paraphrase dataset. Formally, $A = [a_1, \ldots, a_m]$ and $B = [b_1, \ldots, b_m]$ are now of equal size $m$ consisting of documents (sentences) $\{a_i\}$ and $\{b_i\}$. Paraphrases have the same index $i$; letting the symmetric operator $a \sim b$ indicate that $a$ and $b$ are paraphrases of each other, we thus have $a_i \sim b_i$, $i = 1, \ldots, m$. Furthermore we require that each $a_i \in A$ and $b_j \in B$ are unique, that is, there are no duplicates in each source corpus. We can make the additional assumption that $a_i \sim b_j$ iff $i = j$, that is no other inter-corpus paraphrase pairs exist. This, however, is not a formal requirement, but rather a logical consequence of paired paraphrase construction, that even if, say, $a_i$ is *similar* to some $b_j$, $i \neq j$, they are less similar than either $(a_i, b_i)$ or $(a_j, b_j)$. A good embedding space $\mathcal{V}$ and document distance metric $\delta$ will ensure that, with few exceptions, $\delta(a_i, b_j) > \delta(a_i, b_i)$ and $\delta(a_i, b_j) > \delta(a_j, b_j)$ for any $i \neq j$.

| | Corpus A | Corpus B | |
|---|---|---|---|
| $a_1$ | How can I be a good geologist? | What should I do to be a great geologist? | $b_1$ |
| $a_2$ | What is the best book ever made? | What is the most important book you have ever read? | $b_2$ |
| $a_3$ | Can I recover my email if I forgot the password? | What should I do if I forgot my email password? | $b_3$ |
| ... | | | ... |
| $a_n$ | Does a black hole have mass? | Does a black hole have a finite mass? | $b_n$ |

**Figure 6**: Example of paraphrase corpora $A$ and $B$ consisting of two columns of the Quora paraphrase dataset. Texts $a_i \in A$, $b_i \in B$, with the same index $i$, are paraphrases, denoted $a_i \sim b_i$.

Large databases of such paraphrase pairs (e.g, PPDB, [4]) are publicly available, and paraphrase identification and scoring are established machine learning tasks in the natural language processing field. Here, we will not attempt paraphrase identification, but rather use such corpora as an intermediate resource for evaluating metric distributionality.

Consider now a sample $KSC(A, B)$ of length $K + 1$, where $A$ and $B$ are paraphrase corpora. As opposed to the case in Section 3, where $A$ and $B$ were fundamentally different, and hence the distances $D_\ell(A, B, d)$ were expected to increase on average with increasing separation level $\ell$, here there is the balancing effect of semantic similarity from the paraphrase. For instance, $c_0$ and $c_K$ have the highest KSC separation level $\ell = K$, but because they are entirely sampled from $A$ and $B$, respectively, because $A$ and $B$ are paraphrase corpora, there should be a higher level of similarity between $c_0$ and $c_K$ because of the one-to-one matching, than otherwise.

To form a KSC corpora set, we first create sources $A$ and $B$ of equal size $n$ by sampling $n$ paraphrase pairs without replacement from the full dataset.

Figure 7 shows a conceptual visualization of comparisons between KSC corpora $c_i$ formed by double lottery. Here, $K = 5$, and each paraphrase corpus $A$ $(= c_0)$ and $B$ (oranges, $= c_5$) are also of size $n = 5$. On the left, the $n = 5$ blue dots each represent a document $a_j \in A$. Each blue dot has a corresponding document $b_j \in B$, its paraphrase, represented by an orange dot. The dots are plotted in a semantic embedding space, because the paraphrases pair documents are closer to each other than they are to the other documents in either corpus. The blue and orange ovals show visually that the two corpus distributions overlap to a high degree, because they are semantic paraphrases. The center panel shows two pairs of KSC corpora $(c_2, c_3)$ and $(c_1, c_4)$; in each case, each observed $c_i$ (formed by double lottery) is only one possible such sampling that contains $i$ distinct (orange) points from $B$. It shows that $c_2$ and $c_3$ exactly share three points (two orange, one blue), and $c_2$ and $c_5$ exactly share two points (one each orange and blue). On average, sampled $c_2$ and $c_3$ should be closer than the other pair because their $\ell$-separation is 1, while the other pair are $\ell = 3$ separated; however, when $\ell$ increases toward the maximum ($K = 5$), the similarity should increase as well because the two sampled corpora tend to be entirely from one source corpus or the other.

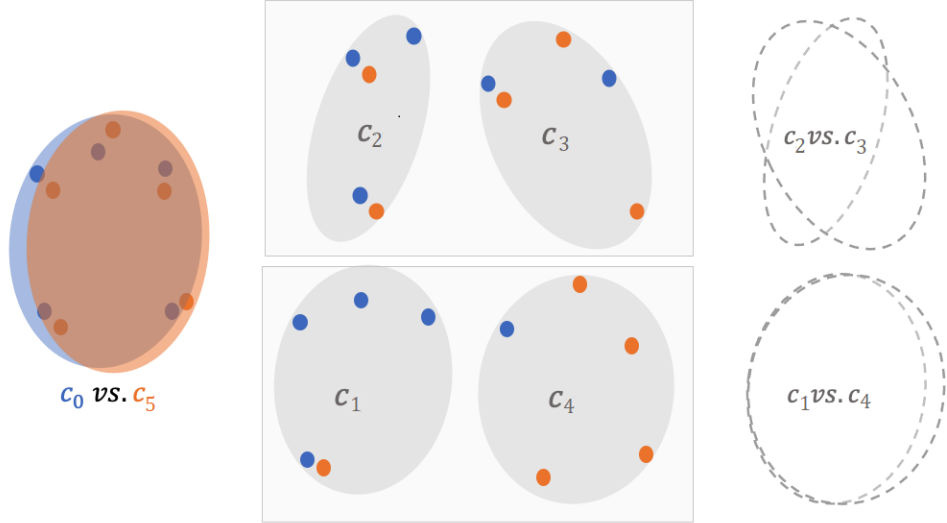

**Figure 7**: Illustration of sampling of KSC corpora from paraphrasesources $A$ (blue) and $B$ (orange).

Figure 8 shows the results of repeating the KSC simulations from [6] except now with $A$ and $B$ being the sampled paraphrase corpora rather than corpora from different domains. The same initial sample of $n = 50$ paraphrase pairs to form $A$ and $B$ is used in all plots. A KSC-set with $K = 15$ is formed $R = 5$ times, yielding $KSC(A, B)^{(r)} = \{c_0^{(r)}, \ldots, c_K^{(r)}\}$ for each of $r = 1, \ldots, R$. Let $V \mathrel{+\!\!+} W$ represent the concatenation (not union, which discards duplicate elements) of two sets $V$ and $W$, and $\mathrel{+\!\!+}_{i=1}^{n} V_i$ represent the concatenation of multiple sets $V_1, \ldots, V_n$. Then, for each distance metric $d$, we plot the observed distribution of the distance values between $\ell$-distant corpora pairs, pooled ($^p$) across sampling repetitions, denoted $D_\ell^p(A, B, d) \mathrel{+\!\!+}_{r=1}^{R} D_\ell^{(r)}(A, B, d)$, versus $\ell = 1, \ldots, K$; see Equation 2.

Figure 8 shows that when the KSC set is constructed from source paraphrase corpora, the $\ell$-distant corpora distances do not simply increase with $\ell$, as in [6], but rather tend to increase first and then decrease towards 0, because the maximal-$\ell$ distant corpora $c_0$ and $c_K$ are the source paraphrase corpora themselves, and thus should have low semantic distance. Furthermore, the distance metrics the intuitively should be more non-distributional (particularly IRPR, $DC_2$ and $PR_2$), tend to have distance values that change significantly over $\ell$; see Section 2 and compare with, say, Figure 2.

Ultimately, we would like to see if we can measure the degree to which a corpus distance $d$ behaves 'distributionally' or not. This is discussed in Section 5.

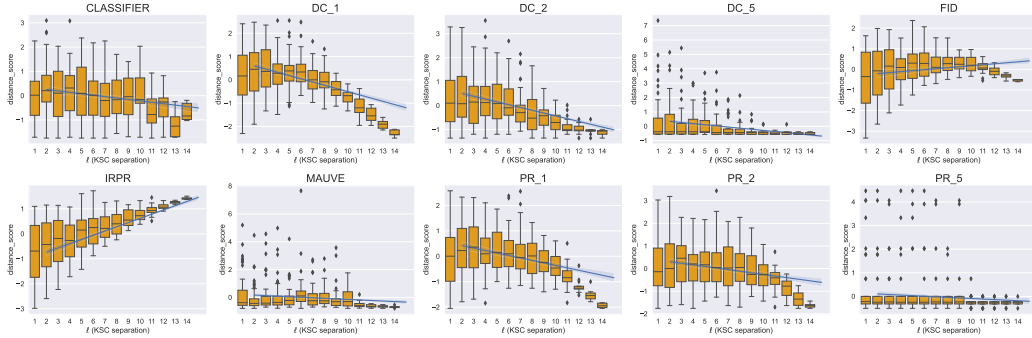

**Figure 8**: Distribution of pooled $\ell$-distant corpora distances $d(c_i, c_j)$, between paraphrase corpora $A$ and $B$. As $\ell$ increases, the distances tend to first increase and then decrease to 0, rather than increase the whole way, since $A$ and $B$ are paraphrase corpora and not from different domain as in [6].

## 5. Evaluating distributionality

Figures 2 and 8 give intuition about the shapes of the trend of $d(c_i, c_j)$ vs $\ell$ for corpus distance metrics $d$ when we know that the underlying mechanism of the distance metric should make it behave more distributionally or not. We propose here that the shape of the trajectories be used as a criterion. Specifically, we will propose two distance metrics which we take as being prototypical or representative of distributionality and non-distributionality, and we will compare other candidate metrics $d$ to them, as follows:

- **Distributional**: Energy distance statistic
- **Non-Distributional**: Average Directed Hausdorff Distance (AHD)

### 5.1 Energy distance

**Energy distance** ([11]) is a statistical distance between two probability distributions, which involves a calculation of the expected distance between independent vectors from two distributions. Here, we use the goodness-of-fit energy statistic, denoted $\mathcal{E}$ (Equation 3), which uses the observed values from two samples. For a given docuement-level distance metric $\delta$ (we use cosine), $\mathcal{E}$ is a calculated as twice the average distance $\delta$ between document pairs in different corpora, minus the average within-corpora document distances. It is thus high (with unbounded maximum) when two corpora $A$ and $B$ are far from each other and the within-corpus distances are small.

$$\mathcal{E}_\delta(A, B) = \frac{2}{|A||B|} \sum_{i=1}^{|A|} \sum_{j=1}^{|B|} \delta(a_i, b_j) - \frac{1}{|A|^2} \sum_{i=1}^{|A|} \sum_{j=1}^{|B|} \delta(a_i, a_j) - \frac{1}{|B|^2} \sum_{i=1}^{|A|} \sum_{j=1}^{|B|} \delta(b_i, b_j) \quad (3)$$

Because the energy distance calculates corpus distances by looking at all pairwise distances, we consider it to be a prototypical distributional metric.

### 5.2 Average Hausdorff Distance (AHD)

Given two finite point sets $X, Y$, and a (possibly asymmetric) between-elements (here, documents) distance metric $\delta$, define $d_h(x, Y) = \inf_{y \in Y} \delta(x, y)$ as the shortest distance $\delta$ between $x$ and any element in $y \in Y$, that is, the nearest-neighbor distance between $x$ and $Y$. The Hausdorff Distance is then defined as $d_H(X,Y) = \max \left( \sup_{x \in X} d_h(x, Y), \sup_{y \in Y} d_h(y, X) \right)$. That is, considering each point set as a starting point, it first finds the largest such nearest-neighbor distance from it to the other. This is the worst-case (largest) distance needed to travel from an element of the point set in that direction. The maximum is then taken because $\delta$ need not be symmetric, to represent the best worst-case bi-directional travel distance from a point in either of the two sets to the other. The Average Hausdorff Distance[4] (AHD) is an adaptation that weights $X$ and $Y$—which may be of differing sizes— equally in calculating the directed Hausdorff distances, and is defined (Equation 4) as

$$d_{AHD} = \left( \frac{1}{|X|} \sum_{x \in X} d_h(x, Y) + \frac{1}{|Y|} \sum_{y \in Y} d_h(y, X) \right) / 2 \quad (4)$$

The AHD $d_{AHD} \in [0, 1]$ is our prototypical *non-distributional distance*, because it is constructed of local nearest-neighbor distances, which neglect larger distributional considerations. It is possible for each member of one set to have a very close, possibly identical, neighbor in the second, yet the overall mass of the point sets may be far from each other. It may be non-robust to small but specific changes to nearest neighbors that have little effect on the overall point masses.

We note that $d_{AHD}$ is very similar to IRPR, in which

$$\text{precision} = \frac{1}{|X|} \sum_{x \in X} d_h(x, Y), \quad \text{recall} = \frac{1}{|Y|} \sum_{y \in Y} d_h(y, X)$$

IRPR is then the harmonic mean of precision and recall, rather than the arithmetic mean, as in $d_{AHD}$.

Figure 9 illustrates calculation of $d_{AHD}(X, Y)$ for two synthetic 2-D numeric samples. Euclidean distance is used as the underlying distance metric $\delta$ for defining neighbors. The grey (solid or dashed) arrows connect a member of the first sample $X$ (hollow green circles) with its closest neighbor in the second sample $Y$ (solid green circles); the solid green arrows are the nearest-neighbor relations in the reverse direction. Because the distance $d_h$ is calculated separately in each of the two directions between the corpora, if $x_i \in X$ is the nearest neighbor in $X$ to some $y_j \in Y$, but the reverse may not be true; and also a given element may be the nearest neighbor of multiple other elements. AHD is the average of the corpus averages of $d_h$ in each direction, as shown in the figure's title. When applied in the text corpora settings, as in energy distance (Section 5.1), cosine distance will be used as the underlying document-level distance metric $\delta$, applied on the vector embedding discussed in Section 4.

In the following Section 5.3, we will see that IRPR and AHD both tend to increase on average with $\ell$, unlike the others, which tend to be lowest on average when $\ell = K$, its highest, that is, when the two compared $c_0, c_K$ are each entirely from the two sources $A, B$ and are paired paraphrase corpora. In this case, $d(c_0, c_K)$ should be relatively small because each item $a_i \in c_0$ has its paraphrase $b_i \in c_K$, which should be very semantically close to it; that is, $\delta(a_i, b_i) > 0$ but is small. Recall, though, that $c_0$ and $c_K$ will not share documents that are exactly the same.

---
[4] As defined in [12] and associated code [2]; see also [1], equation 1, for clarifying notation and discussion of its usage in image segmentation.

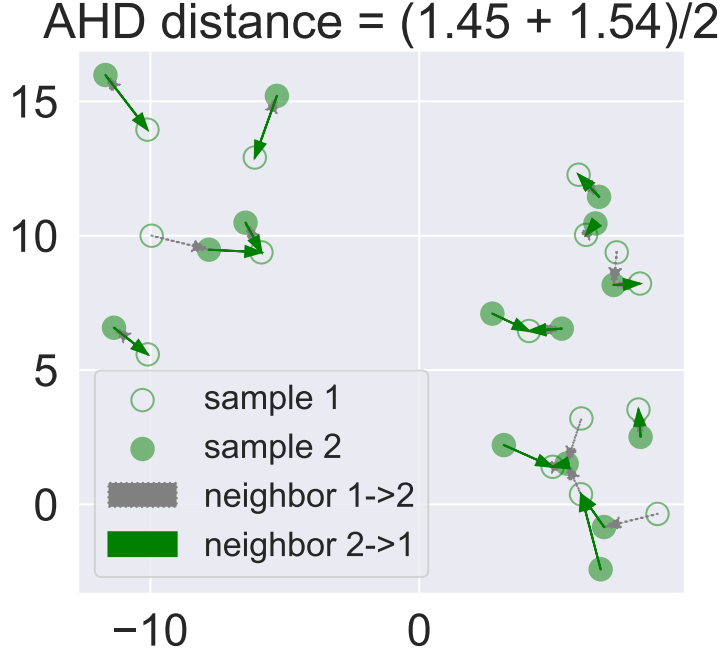

**Figure 9**: Illustration of AHD between two synthetic 2-D point sets.

However, when $\ell < K$, the compared $c_i$ and $c_j$ may actually share documents that are exactly the same (and therefore will have exactly 0 distance) due to the double lottery process. For instance, if $a_v \in c_i$ is actually $= b_w \in c_j$ (the same document $a_v = b_w$ was sampled in both corpora), then $d_h(a_v, c_j) = d_h(b_w, c_i) = 0$. When this happens enough, the precision and recall, which average these directional distances, will tend to decrease. However, when $c_0, c_K$ are compared, there will be no such cases $a_v \in c_0$ where $d_h(a_v, c_K) = 0$ since the closest element in $c_K$ will tend to be its paraphrase $b_v \in c_K$, which has nonzero distance from it. Therefore, the IRPR and AHD distance when $d(c_0, c_K)$ will actually tend to be high, rather than low, relative to other pairs $(c_i, c_j)$ for which $j - i < \ell$. This aspect reflects the non-distributional aspects of the distance metrics IRPR and AHD, that they are affected by (relatively small) document-level distances between paraphrases, when in fact the two compared corpora are most similar in overall distribution.

### 5.3 Standardized comparison of KSC patterns

Our determination of a metric $d$'s distributionality will be based on the similarity of its KSC distances $D_\ell(A, B, d)$ vs $\ell$ to $D_\ell(A, B, d^*)$, where $d^*$ is either of the prototypical distances $\mathcal{E}_\delta$ or $d_{AHD}$. Note that in Figure 8, because the various metrics $d$ may have differing typical raw ranges of values, we cannot use the raw values for comparison because we are interested in the shape of the trajectory, regardless of the raw values. Therefore, for each $d$, we perform standardization as such: let $P_d = ++_{r=1}^{R} ++_{\ell=1}^{K} D_\ell^{(r)}(A, B, d)$ be the values of pairwise KSC corpora distances, pooled across separations $\ell$ and repetitions $r$ of the KSC sampling. Letting $\mu_S$ and $\sigma_S$ be the sample mean and standard deviation of the pooled $P_d$, let $\tilde{D}_\ell^p(A, B, d)$ be the values of $D_\ell^p(A, B, d)$ with $\mu_S$ subtracted and $\sigma_S$ divided, so that when pooled across $\ell$, the mean is now 0 and standard deviation is 1. Figure 10 shows the re-sult of this standardization on the metrics of interest (top), which correspond to the plots in Figure 8. The bottom panel shows the standardized plots for the two prototypical distances, for comparison.

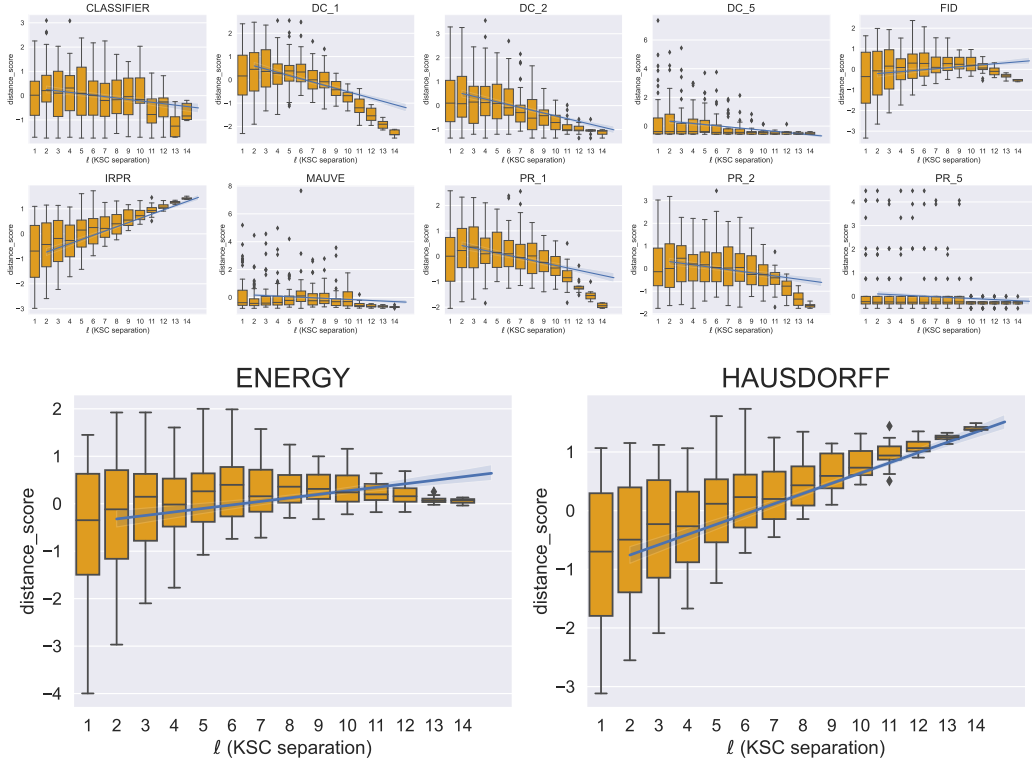

**Figure 10**: Standardized KSC distance plots for a fixed sample.
Top: The metrics to be evaluated, corresponding to Figure 8.
Bottom: The two prototypical distance metrics.

We now propose to conduct the comparisons via kernel density estimate (KDE) distances. Let $\hat{f}_d^\ell$ be a kernel density function estimate of the repetition-pooled sample $\tilde{D}_\ell^p(A, B, d)$. We adapt Fan's ([3]; see [7]) squared deviation between two estimated density functions $\hat{f}$ and $\hat{g}$, defined as

$$I_{f,g} = \int_{-\infty}^{\infty} \left( \hat{f}(x) - \hat{g}(x) \right)^2 \mathrm{d}x$$

by performing a discrete approximation between endpoints $(a, b) = (-8, 8)$ (since the distances are standardized) at $c = 3000$ equally-spaced points with bin width $\delta_x = (b - a)/c$. For two distance metrics $v, w$, the $\ell$-conditional density distance is

$$\hat{I}_{v,w}^\ell = \sum_{i=0}^{c} \left( \hat{f}_v^\ell(a + i\delta_x) - \hat{f}_w^\ell(a + i\delta_x) \right)^2 \delta_x$$

To calculate the overall distance between the KSC trajectories of the two metrics, across separations $\ell$, we weight these $\ell$-conditional distances by the sample size. That is, let $w_\ell = \frac{|\tilde{D}_\ell^p(A,B,d)|}{|P_d|}$, and let

$$\hat{I}_{v,w} = \sum_{\ell=1}^{K} w_\ell \hat{I}_{v,w}^\ell$$

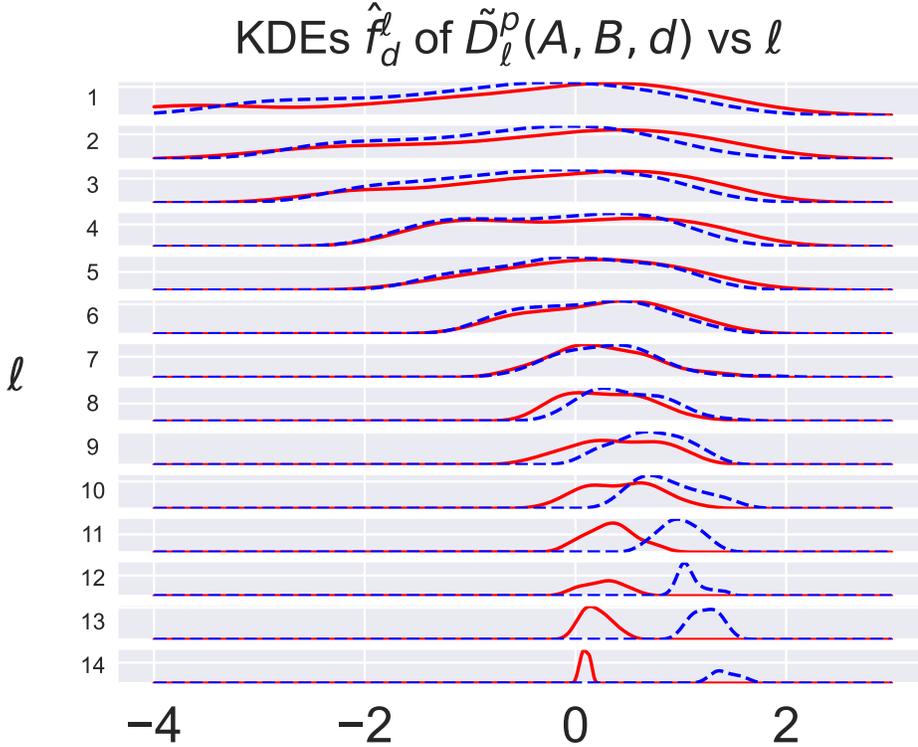

**Figure 11**: The red solid lines correspond to ENERGY, the dashed blue to HAUSDORFF.

The procedure to classify a given distance metric $d$ can thus be to calculate $\hat{I}_{d,d^*}$ for $d^* \in \{\text{ENERGY}, \text{AHD}\}$, and to assign $d$ to belong to the type $d^*$ with lowest distance $\hat{I}_{d,d^*}$.

Figure 11 shows, for each $\ell$, the KDE $\hat{f}_d^\ell$ of the two prototypical distances (Figure 10, bottom). The KDEs and the trajectories of the plots differ most when $\ell$ is high. As noted in Section 5.2, this is caused by the fact that that when $\ell$ is maximized, the pair $c_0, c_K$, which consist entirely of paraphrase document pairs. For AHD, these close pairs have nonzero document corpus distanc, causing the corpus distance $d$ to rise, while under other metrics $d$, these paraphrase corpora tend to have *less* distance than other pairs.

Figure 12 shows the values of the deviations $\hat{I}_{d,d^*}$ for each of the candidate metrics $d$. We see that only for $d = \text{IRPR}$ is $\hat{I}_{d,ENERGY} > \hat{I}_{d,AHD}$ (confirming that it is more non-distributional), while for the others the reverse is true. We do see, however, that for many, particularly the DC and PR-based metrics, the similarity to *either* of the baselines is not very strong. This is analyzed further in Section 5.4.

### 5.4 DC and PR-distance components

As noted in [6], the DC (density-coverage) and PR (precision-recall) distance metrics are calculated as 1 minus the harmonic mean of the two components—density and coverage, or precision and recall—when a fixed number of $k$-nearest neighbors is used for the calculation. Our initial assumption was that higher $k$ (perhaps up to a limit) indicated more distributionality, which does not seem to be evident in practice in the textual corpora, in contrast to that observed in Section 2 for synthetic numeric samples. Here, we investigate patterns in the components of these distances over different $\ell$ for values of $k \in \{1, 2, 3, 5, 10\}$.

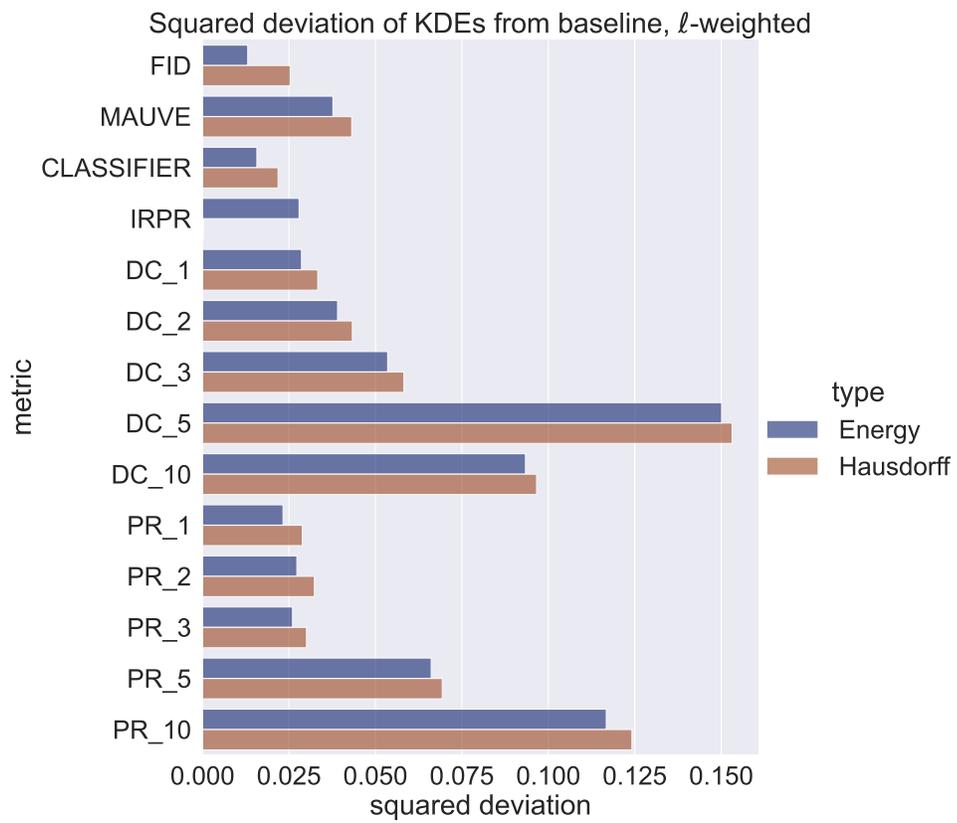

Figure 12: Deviation $\hat{I}_{d,d^*}$ for candidate metrics $d$ versus $d^* \in \{\text{ENERGY}, \text{AHD}\}$.

Figure 13 shows boxplots of these component values of the raw $D_\ell^p(A, B, d)$. The colors represent different choices of $k$ neighbors. As noted in [6], the PR and DC distance metrics are motivated by applications in computer vision where there is a target and reference sample. The components are each defined in one direction between two corpora; PR is a symmetric metric because the two components are equivalently-defined if the directions are switched, but DC's components are not, hence it is not symmetric. When comparing corpora $c_i, c_j$ with a document-level distance metric $\delta$ (we use cosine), and a document's $k$-nearest neighbors in the other corpus, the components are defined as follows:

- **Precision** is the average proportion of elements in $c_j$ that are $k$-nearest among $c_j$ to any element in $c_i$; **recall** is the same, in the reverse direction, for $c_i \to c_j$.

- **Density** is the the average number of elements in $c_i$ that an element in $c_j$ is $k$-nearest to, divided by $k$.

- **Coverage** is the proportion of elements in $c_i$ whose nearest neighbor in $c_i$ is closer to it than is its $k^{\text{th}}$ nearest in $c_j$.

The precision and recall should both be high (near 1) when $c_i, c_j$ share many paraphrase pairs, and hence the PR distance should be small (close to 0). Figure 13 shows that the variability in precision, recall, and coverage decrease sharply as the number of neighbors $k$ increases; for $k \in \{1, 2, 3\}$ there are observable differences across $\ell$ and within these components for the compared $\ell$-distant corpora pairs $KSC_\ell(A, B)$, but for higher $k > 3$ when $c_i, c_j$ contain documents that are paraphrases of each other, all three of these components become nearly 1. For instance, precision with $k = 1$ measures the proportion of elements in $c_j$ that are closest among $c_j$ to at least one element in $c_i$. If many elements in $c_j$ have a paraphrase (or duplicate) in $c_i$, that they would be that element's closest neighbor in $c_i$, the precision would be relatively high, but still variable depending on the degree of overlap; the same applies to recall. But when $k$ is too high, then the nearest-neighborhood radii get very large, and nearly every element in $c_j$ or $c_i$ should be in one of them, leading the precision, recall, and coverage to be nearly 1 and erasing the distinction in distances we should see due to changes in $\ell$. Again, this is true when $A$ and $B$ are paraphrase corpora and the sampled KSC corpora have size equal to $A$ and $B$; it would not be the case if the the KSC corpora were sampled in sizes (i.e. samples small relative to the size of the sources) that the chance of overlaps was negligible. With density, there is again decreasing variability when $k$ increases (the boxplots shrink to 1), but only for the first several $\ell$. After $\ell \geq 6$, the densities are nearly all 1 for all $k$. Thus, we see that $k$ cannot be too large if the PR and DC metrics are to be informative in this setting. Thus, it is not necessarily that higher $k$ indicates better ability to distinguish global distributional differences.

## 6. Conclusions

In this work, we introduced the concept of distributionality, defined as the sensitivity of global a two-sample distance metric to small perturbations in individual sample elements. This concept was applied to text corpus distance metrics from [6] on Known Similarity Corpora (KSCs) of paraphrases. The degree of distributionality was determined by comparison of the metric's measured trajectories versus those of two prototypical metrics.

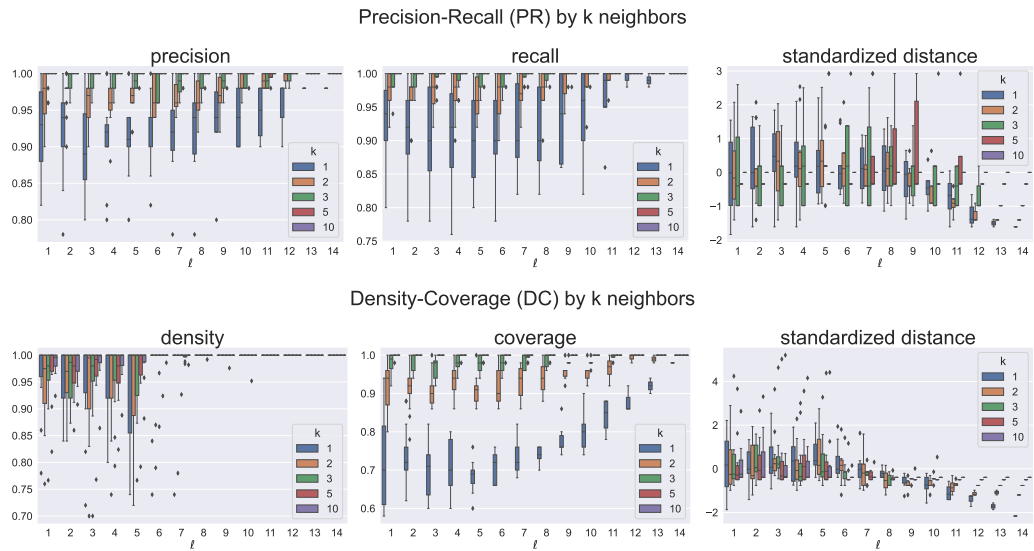

**Figure 13**